\documentclass[final]{cvpr}

\usepackage{times}
\usepackage{epsfig}
\usepackage{graphicx}
\usepackage{amsmath}
\usepackage{amssymb}

\usepackage{nicefrac}       
\usepackage{microtype}      
\usepackage{balance}

\usepackage{multirow}
\usepackage{verbatim}
\usepackage{color}
\usepackage{float}
\usepackage{enumitem}
\usepackage{booktabs}
\usepackage{tabulary,multirow,overpic,xcolor}

\usepackage[caption=false]{subfig}
\usepackage{pifont}
\usepackage{epstopdf}
\usepackage{algorithm}
\usepackage{algorithmicx}
\usepackage{algpseudocode}
\usepackage{enumerate}

\usepackage{bm}
\usepackage{t1enc}
\usepackage{colortbl}
\usepackage{cite}

\usepackage[british,UKenglish,USenglish,english,american]{babel}

\newcommand{\fref}[1]{Fig.~\ref{#1}}
\newcommand{\tref}[1]{Table~\ref{#1}}
\newcommand{\sref}[1]{Sec.~\ref{#1}}

\usepackage{soul}
\definecolor{carmine}{rgb}{0.59, 0.0, 0.09}

\newcommand{\app}{\raise.17ex\hbox{$\scriptstyle\sim$}}

\def\x{$\times$}

\newcolumntype{x}[1]{>{\centering\arraybackslash}p{#1pt}}

\newlength\savewidth\newcommand\shline{\noalign{\global\savewidth\arrayrulewidth
		\global\arrayrulewidth 1pt}\hline\noalign{\global\arrayrulewidth\savewidth}}
\newcommand{\tablestyle}[2]{\setlength{\tabcolsep}{#1}\renewcommand{\arraystretch}{#2}\centering\footnotesize}

\makeatletter\renewcommand\paragraph{\@startsection{paragraph}{4}{\z@}
	{.5em \@plus1ex \@minus.2ex}{-.5em}{\normalfont\normalsize\bfseries}}\makeatother

\definecolor{fastcolor}{RGB}{100,178,100}
\definecolor{slowcolor}{RGB}{120,120,243}
\definecolor{expandcolor}{RGB}{244,157,78}
\definecolor{clipscolor}{HTML}{0071bc}

\definecolor{predictioncolor}{RGB}{0,255,0}
\definecolor{labelcolor}{RGB}{255,0,0}

\definecolor{demphcolor}{RGB}{144,144,144}

\definecolor{xycolor}{RGB}{60, 120, 216}

\definecolor{wcolor}{RGB}{103, 78, 167}

\definecolor{dcolor}{RGB}{166, 77,21}

\definecolor{gcolor}{RGB}{204, 102, 153}

\definecolor{tcolor}{RGB}{80, 200, 180}

\definecolor{eicolor}{RGB}{153, 51, 102}

\definecolor{gridcolor}{RGB}{0,94,255}
\newcommand{\gridcolor}[1]{\textcolor{gridcolor}{#1}}

\newcommand{\blocket}[4]{\multirow{3}{*}{\(\left[\begin{array}{c}\text{1$\times$1$^\text{2}$, #1}\\[-.1em] \text{$3$$\times$3$^\text{2}$, #2}\\[-.1em] \text{1$\times$1$^\text{2}$, #3}\end{array}\right]\)$\times$#4}
}

\usepackage[pagebackref=true,breaklinks=true,colorlinks,bookmarks=false]{hyperref}

\def\cvprPaperID{4949} 
\def\confYear{CVPR 2021}

\begin{document}

\title{Coarse-Fine Networks for Temporal Activity Detection in Videos \vspace{-6mm}}

\author{Kumara Kahatapitiya and Michael S. Ryoo\\
Stony Brook University, Stony Brook, NY 11794, USA\\
{\tt\small \{kkahatapitiy,mryoo\}@cs.stonybrook.edu}\vspace{-4mm}
}

\maketitle

\begin{abstract}
    \vspace{-2mm}
   In this paper, we introduce \emph{Coarse-Fine Networks}, a two-stream architecture which benefits from different abstractions of temporal resolution to learn better video representations for long-term motion. Traditional Video models process inputs at one (or few) fixed temporal resolution without any dynamic frame selection. However, we argue that, processing multiple temporal resolutions of the input and doing so dynamically by learning to estimate the importance of each frame can largely improve video representations, specially in the domain of temporal activity localization. To this end, we propose (1) `Grid Pool', a learned temporal downsampling layer to extract coarse features, and, (2) `Multi-stage Fusion', a spatio-temporal attention mechanism to fuse a fine-grained context with the coarse features. We show that our method outperforms the state-of-the-arts for action detection in public datasets including Charades with a significantly reduced compute and memory footprint. The code is available at \href{https://github.com/kkahatapitiya/Coarse-Fine-Networks}{https://github.com/kkahatapitiya/Coarse-Fine-Networks}.
   \vspace{-6mm}
\end{abstract}

\section{Introduction}
\vspace{-1mm}




Learning to represent videos is important. It requires embedding both spatial and temporal information in a sequence of frames, often implemented with 3D convolutions. Leaning to build good video representations is crucial for various vision tasks including action classification, video object segmentation, and complex human activity recognition as well as temporal localization of such activities.


One of the main challenges in video representation learning is in capturing long-term motion from a continuous video. 
In order for a convolutional neural network to abstract long-term motion information across many frames, a large number of (spatio-)temporal conv. layers (or such layers with large kernels) are necessary, requiring many parameters. This, combined with the difficulty in obtaining large-scale annotated videos and increased computation time, makes the learning of the video representation very challenging for non-atomic activities. This is even more challenging for temporal activity detection (i.e., localization), as the activities may very often temporally overlap.
A mechanism to reliably and efficiently capture various motion in videos is necessary.

\begin{figure}[t]
	\centering
	\hspace{-10mm}
	\vspace{-1mm}
	\includegraphics[width=1.\linewidth]{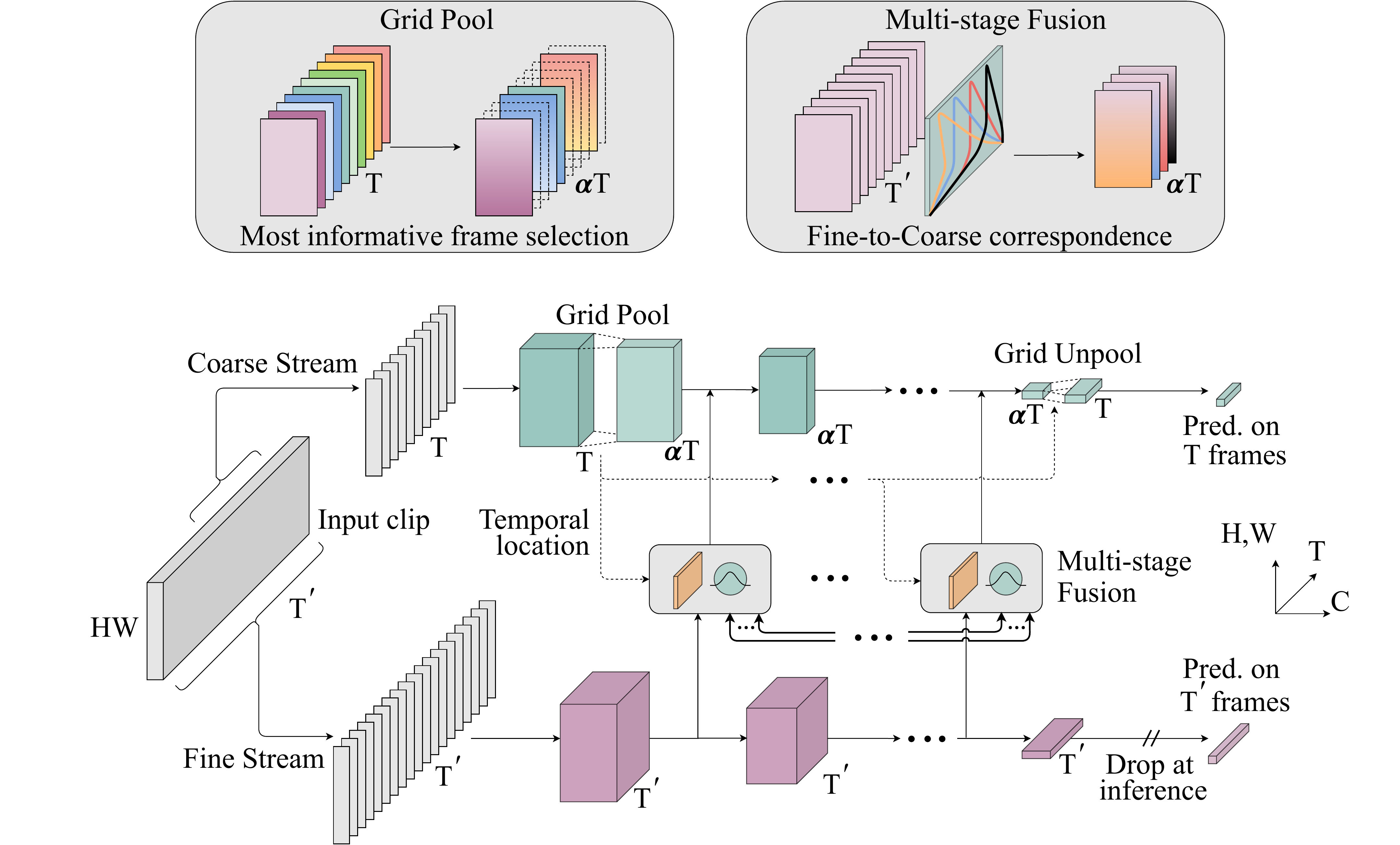}
	\caption{\textbf{Coarse-Fine Networks} process information at two different temporal resolutions. The Coarse stream learns to sample the most informative frame locations through a learnable downsampling operation: \textit{Grid Pool}, whereas the Fine stream process the entire temporal duration of the input to extract a fine-grained context. The connections in-between the two streams: \textit{Multi-stage Fusion}, provide multiple abstraction-levels of the fine-grained context, calibrated to the temporal locations of the coarse frames. For Charades dataset \cite{sigurdsson2016hollywood}, we configure our network to use $T=64$, $T^{'}=128$ and $\alpha=1/4$.}
	\vspace{-5mm}
	\label{fig:intro}
\end{figure}

Use of frame striding or temporal pooling (i.e., lowering the frame rate) has been a successful strategy to cover a larger time interval without increasing the number of model parameters. Since such striding loses fine details of frame changes, it was often paired with another CNN tower taking an input with a higher frame rate, forming a two-stream (or multi-stream) CNN architecture as was done in SlowFast \cite{feichtenhofer2019slowfast} and AssembleNet \cite{ryoo2019assemblenet}. 
These models confirmed the benefits of frame striding as well as multi-stream architectures to combine representations with multiple temporal resolutions.

However, although using temporal striding (with a multi-stream multi-resolution architecture) allows the model to more easily process long-term motion, they are limited as it ignores `importance' of each frame.
Informativeness of each frame is different. It is often unnecessary and redundant to feed almost identical frames as an input to the model when there is no/little motion in video frames. On the other hand, if a human in the video is displaying a rapid motion, taking all such frames into consideration is desired. Uniform temporal striding or pooling is incapable of such dynamic frame selection. 

\begin{figure}[t]
	\centering
	\includegraphics[width=0.9\linewidth]{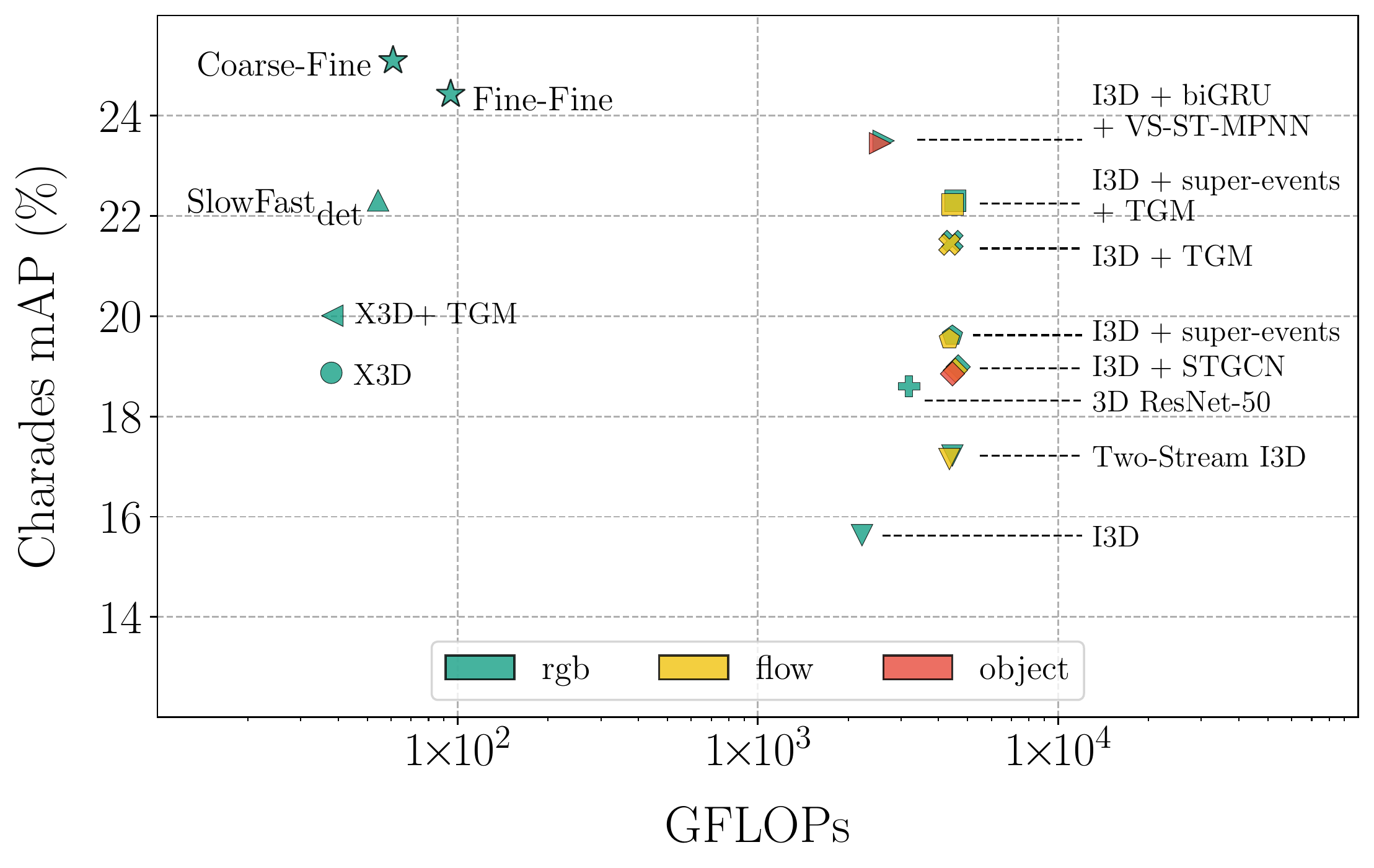}
	\caption{\textbf{Performance/complexity trade-off} of state-of-the-art methods for activity `localization' in Charades \cite{sigurdsson2016hollywood}. Our Coarse-Fine Networks achieve superior performance than the previous best-performing method in literature, with more than one order of magnitude reduction in compute. Moreover, we do not use any additional modalities such as optical flow or object detections.}
	\label{fig:performance_complexity}
	\vspace{-4mm}
\end{figure}

In this paper, we propose (1) a new approach that allows a learnable dynamic selection of temporal frames within the model, as well as (2) a method to fuse such sampled (i.e., temporally `coarse') representations with conventional, more temporally `fine' representations. We introduce the \emph{Coarse-Fine Networks}. A new component named temporal Grid Pooling is presented to obtain better Coarse representations, and the Multi-stage Fusion is introduced to best combine such Coarse representations with the conventional Fine representations. Unlike \cite{feichtenhofer2019slowfast,ryoo2019assemblenet}, our Grid Pooling learns to dynamically select informative frames.
\fref{fig:intro} illustrates the overview of the model, and \fref{fig:performance_complexity} shows the benefits of the model, which we discuss more in the paper. \vspace{-2mm}


\section{Related Work}

CNNs learning 3D spatio-temporal representations for human activity recognition have been very successful \cite{tran2014c3d, karpathy2014large,carreira2017quo, tran2017convnet, hara2017learning}. Two-stream approaches were often designed to combine RGB and optical flow \cite{simonyan2014two,feichtenhofer2016convolutional}, particularly focusing on video classification. SlowFast network \cite{feichtenhofer2019slowfast} showed the potential that combining representations of different temporal resolutions (i.e., frame rates) could also benefit action recognition. More recently, AssembleNet \cite{ryoo2019assemblenet} showed the effectiveness of multi-stream models with neural architecture search, and X3D \cite{feichtenhofer2020x3d} studied computationally more efficient 3D conv. modules. There also are approaches focusing on the modeling of temporal structure in videos, often particularly designed to handle longer videos (with long-term motion) \cite{yue2015beyond,varol2017long,lea2017temporal,piergiovanni2018learning,piergiovanni2019temporal}. Another group of approaches took advantage of graph representations to model human/object dynamics in the videos, often paired with sequential models \cite{ghosh2020stacked,mavroudi2020representation,ji2020action}. Approaches to explicitly take advantage of objects in videos have been studied as well \cite{ma2018attend,baradel2018object,zhou2019grounded}.

\vspace{-3pt}
\paragraph{Action localization:} There are also a line of work focusing on the temporal action localization task. In the localization task (e.g., Charades localization \cite{sigurdsson2016hollywood}), the objective is not about making a classification decision per segmented video but about annotating every frame with multiple ongoing activities. Use of sequential models such as LSTMs have been popular \cite{yeung2016end,escorcia2016daps,yeung2018every}, and fully convolutional approaches also showed promising results \cite{shou2016temporal,zhao2017temporal,xu2017r,shou2017cdc}.

\vspace{-3pt}
\paragraph{Dynamic sampling:} Selective processing of information has been of interest to the computer vision community. From Deformable convolutions \cite{dai2017deformable} to Graphical networks \cite{scarselli2008graph, liu2017dynamic, xinyi2018capsule}, various core components of neural networks are based on this idea. Multiple recent works also try to address dynamic sampling of inputs, either spatially \cite{recasens2018learning, jaderberg2015spatial, gao2020beyond}, temporally \cite{zolfaghari2018eco, wu2019adaframe, korbar2019scsampler, wu2019liteeval} or spatio-temporally \cite{meng2020ar}. 

\vspace{-2mm}
\section{Coarse-Fine Networks}
\vspace{-2mm}

Coarse-Fine Networks explore how video architectures can benefit from different abstractions of temporal resolution and long-term temporal information. As shown in \fref{fig:intro}, we do this by processing the information at two different temporal resolutions: coarse and fine, in a two-stream architecture. The Coarse stream learns to (differentiably) select the most informative frame locations, essentially performing a learned temporal downsampling to abstract a lower temporal resolution. In contrast, the Fine stream processes the input at the original temporal resolution and provide a fine-grained context to the Coarse stream through a fusion mechanism. To abstract this context information, the Fine stream always looks at the full temporal duration of the input clip (which later gets pooled with Gaussians), whereas the Coarse stream can either look at a shorter clip or the entire clip depending on the inference interval.



In Coarse-Fine Networks, we address two key challenges: (i) how to abstract the information at a lower temporal resolution meaningfully, and, (ii) how to utilize the fine-grained context information effectively. First, to abstract coarse information, we propose \textit{Grid Pool} (\sref{subsec:gridpool}), a learnable temporal downsampling operation which adaptively samples the most informative frame locations with a differentiable process.
Secondly, to effectively use the fine-grained context provided by the Fine stream, we introduce \textit{Multi-stage Fusion} (\sref{subsec:fusion}), a set of lateral connections between the Coarse and Fine streams, which looks at multiple abstraction levels of fine-grained information. 

\vspace{-2mm}
\subsection{Grid Pool}
\label{subsec:gridpool}
\vspace{-1mm}

\begin{figure}[t]
	\centering
	\includegraphics[width=0.8\linewidth]{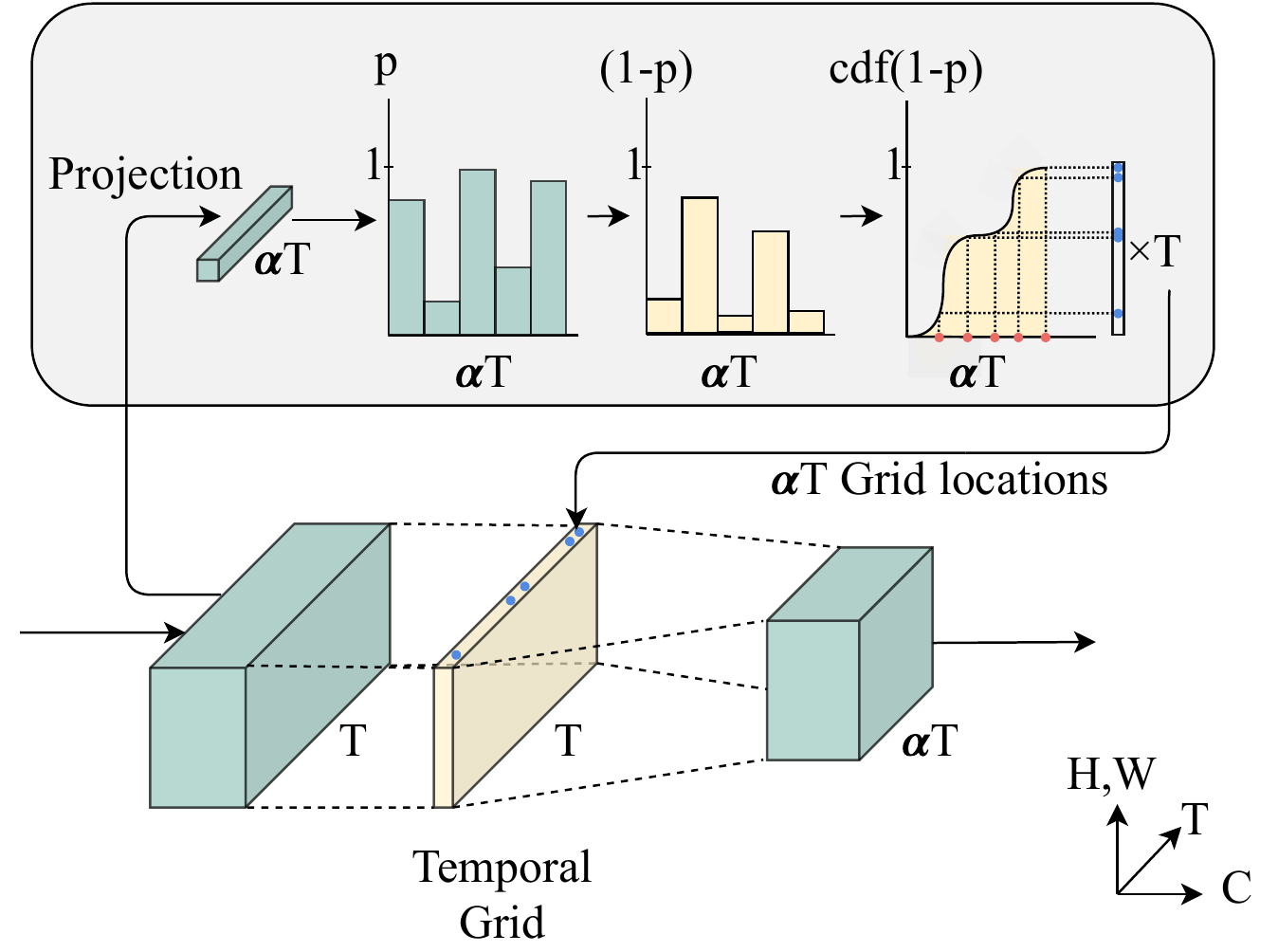}
	\caption{\textbf{Grid Pool} layer learns a temporal downsampling operation based on non-uniform grid locations. $\alpha T$ number of points are differentiablly sampled from an input feature of length $T$ by learning their importance. We interpret $p_{i}$ as the importance of each frame location. Since we want to sample with a lower sampling duration (i.e., a higher frame rate) where we have a higher importance, we construct $\texttt{cdf}(1-p_{i})$ for sampling.} \vspace{-4mm}
	\label{fig:grid}
\end{figure}

Our temporal Grid Pool operation learns the most informative frame locations from a given input clip, and samples the representations corresponding to the locations based on interpolation. In fact, it can be viewed as a learnable temporal downsampling layer with a small compute overhead, which can replace the conventional temporal pooling operations. However, in contrast to these pooling operations, our Grid Pool samples by interpolating on a non-uniform grid with learnable (and adaptive) grid locations as shown in \fref{fig:grid}. First, a lightweight head ($h$) projects the input feature ($\mathbf{X}^C$) of temporal length $T$ to a set of confidence values $\{p_{i}\}_{i=1,\cdots,\alpha T}$, where $\alpha<1$ and $\alpha T$ is an integer (e.g., $\alpha = 1/4$ and $T=128$). These confidence values represent how informative each temporal interval with a size of $1/\alpha$ (e.g., 4 frames if $\alpha = 1/4$) is, and is modeled as a function of the input representation $\mathbf{X}^C$:
\begin{equation}
    \{p_{i}\}_{i=1,\cdots,\alpha T} = h(\mathbf{X}^C)\;.\\ 
\end{equation}

The intuition here is to sample frames at a higher frame rate where the confidence (i.e., informativeness) is high and at a lower frame rate where it is low. In other words, the stride between the interpolated frame locations should be small where the confidence is high, and vice-versa. We normalize these confidence values $p_{i}$ since we need the relative (not absolute) confidence to capture the relative importance of frames. To get a set of $\alpha T$ grid locations based on confidence values, we consider the Cumulative Distribution Function $\{\texttt{cdf}(1-p_{i})\}_{i=1,\cdots,\alpha T}$, which is a non-uniform and monotonically-increasing function. The input of the Grid Pool layer $\textbf{X}^C$ is sampled/interpolated based on these grid locations to get the output $\mathbf{\Tilde{X}}^C$, while making it fully differentiable for backpropagation. This process can be represented as,
\setlength{\jot}{3pt}
\begin{align}
	\nonumber&q_t = T \cdot \texttt{cdf}(1-p_{t}) = T \cdot \frac{\sum_{i=1}^{t}(1-p_{i})}{\sum_{i=1}^{\alpha T}(1-p_{i})}\;, \\ 
	&\mathbf{\Tilde{X}}^C = I\big(\mathbf{X}^C, \{q_t\}_{t=1,\cdots,\alpha T}\big)\;, \label{eq:interp}
\end{align}
where $q_t$ represents the grid location $t$, and $I(\cdot)$ represent the temporal sampling function. Here, when a grid location is non-integer, the corresponding sampled frame is a temporal interpolation between the adjacent frames. We do not perform any spatial sampling in the Grid Pool layer.

\paragraph{Grid Unpooling:}
A temporal interpolation based on a non-uniform grid as such can affect the temporal structure of the propagated features.
Before feberating the final output, the frame-wise predictions of the network should be re-aligned properly for activity detection tasks. To do this, we introduce a \textit{Grid Unpool} operation, which is coupled with the grid locations learned by the Grid Pool layer. This does not have any learnable parameters and simply performs the inverse operation of the former. First, $\alpha T$ re-sample grid locations are calculated based on the inverse mapping of the \texttt{cdf}, based on which, the logits are re-sampled to acquire the original temporal structure. The idea is to re-sample with a low frame-rate in the regions where we used a high frame-rate in Grid Pool, and vice-versa. Any non-integer frame location is temporally interpolated similar to Eq. \ref{eq:interp}. Finally, these logits are uniformly upsampled through interpolation to fit the input temporal resolution. For classification tasks, the Grid Unpool operation may not be necessary as a global pooling of the logits is considered as the prediction.

\subsection{Multi-stage Fusion}
\label{subsec:fusion}

\begin{figure}[t]
	\centering
	\vspace{-1em}
	\includegraphics[width=0.9\linewidth]{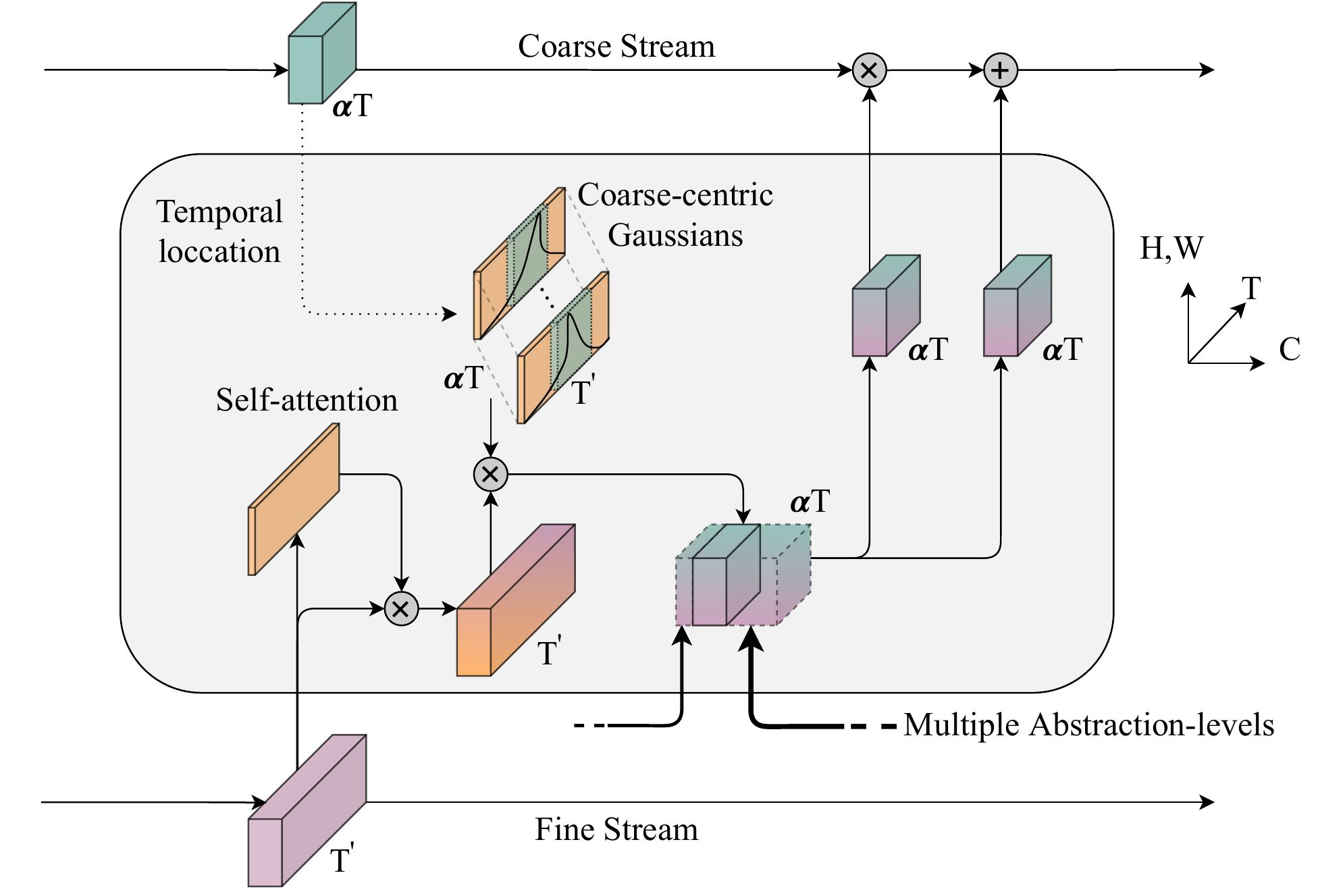}
	\caption{\textbf{Multi-stage Fusion} feeds multiple abstraction-levels of fine-grained context to the Coarse stream. First, Fine stream features get filtered through a self-attention mask. Then, these features get calibrated for each coarse frame, based on Gaussian weights centered at the corresponding coarse frame location. Finally, such calibrated features from multiple abstraction-levels get point-wise convolved to calculate the scale and shift features which provide an affine transformation to the coarse features.}
	\label{fig:fusing}
\end{figure}

We introduce Multi-stage Fusion, a set of lateral connections between the two streams as shown in \fref{fig:fusing}, to fuse the context from the Fine stream with the Coarse stream. We consider three main design choices here: (i) it should be capable of filtering out which fine-grained information should be passed down to the Coarse stream, (ii) it should have a calibration step to properly align the fine features with the coarse features based on their relative temporal locations, and (iii) it should be able to learn and benefit from multiple abstraction-levels of fine-grained context at each fuse-location in the Coarse stream. Our design tries to address these aspects.

\paragraph{Filtering fine-grained information:} First, to decide which fine-grained context should be passed through to the fusion process, the fine feature $\mathbf{X}^F_{l_i}$ at abstraction-level $l_i$ is multiplied with a self-attention mask. This mask is calculated by processing the fine feature through a lightweight head ($g$) consisting of point-wise convolutional layers followed by a sigmoid non-linearity. 
\begin{align}
     \nonumber\mathbf{\bar{X}}^F_{l_i} = \mathbf{X}^F_{l_i} \;g(\mathbf{X}^F_{l_i})
\end{align}

\paragraph{Fine-to-Coarse correspondence:} The attention-weighted fine feature $\mathbf{\bar{X}}^F_{l_i}$ still needs to be calibrated for the temporal location of each coarse feature. Since the Coarse and Fine streams does not necessarily process the same, properly aligned temporal length because of our non-uniform Grid Pooling, 
we need to explicitly compute the frame correspondence. 
To make this calibration, we use a set of temporal Gaussian distributions centered at each coarse frame location $\{\mu^C_{j}\}_{j=1,\cdots,\alpha T}$ which abstract a location-dependent weighted average of the fine feature. We use $\alpha T$ such \textit{Coarse-centric Gaussians}, each having a temporal length of $T^{'}$ and a standard deviation $\sigma$ which is a fraction of this length.
We found that considering the center and scale of these Gaussians to be hyperparameters rather than making them learnable, gives a better performance, possibly due to relatively simpler training. This calibration step can be viewed as,\vspace{-2mm}
\begin{align}
   \nonumber&G^C_j = \frac{1}{\sqrt{2\pi \sigma^2}}\exp{\frac{(t-\mu_j)^2}{2\sigma^2}}\;\;\bigg|_{j=1,\cdots,\alpha T}\;, \\
   &\nonumber\mathbf{\hat{X}}^F_{l_i} = \mathbf{\bar{X}}^F_{l_i} \cdot G^C\;,
\end{align}
where $G^C$ is the stacked Coarse-centric Gaussians and $t\in[0,T^{'}-1]$.

\paragraph{Multiple abstraction-levels:} The feature $\mathbf{\hat{X}}^F_{l_i}$ still corresponds to a single abstraction-level $l_i$ of fine features, where we have Multi-stage Fusion connections in multiple abstraction-levels, i.e., depths of the network. Therefore, we allow each fusion connection to look at the features from all abstraction levels by concatenating them channel-wise (after adjusting the spatial resolution through max pooling), and performing point-wise (i.e. $1\times1\times1$) convolutions to get the final scale ($\mathbf{A}_{l_i}$) and shift ($\mathbf{B}_{l_i}$) features at each fusion location. This can be represented as,
\begin{align}
   &\nonumber\mathbf{\hat{X}}^F = \oplus_{i=1}^{n} \mathbf{\hat{X}}^F_{l_i}\;, \\
   &\nonumber \mathbf{A}_{l_i} = f^{\mathbf{A}}_{l_i}(\mathbf{\hat{X}}^F)\;, \mathbf{B}_{l_i} = f^{\mathbf{B}}_{l_i}(\mathbf{\hat{X}}^F)\;,\\
   \nonumber&\mathbf{\hat{X}}^C_{l_i} = \mathbf{A}_{l_i} \mathbf{\Tilde{X}}^C_{l_i} + \mathbf{B}_{l_i}\;.
\end{align}
where $\oplus$ is the channel-wise concatenation of features from $n$ abstraction-levels and, $f^{\mathbf{A}}_{l_i}$ and $f^{\mathbf{B}}_{l_i}$ represent projection heads to calculate the scale and shift features at each fusion location $l_i$, respectively. This design enables Multi-stage Fusion to process multiple abstraction-levels of fine-grained context through filtering and temporal calibration.

\subsection{Model Details}
\label{subsec:implementation}

\newcolumntype{P}[1]{>{\centering\arraybackslash}p{#1}}
\begin{table}[t]
	\scriptsize
	\centering
	\resizebox{0.8\columnwidth}{!}{
		\tablestyle{1pt}{1.08}
		\begin{tabular}{c|P{1.5cm} P{1.2cm}|P{1.2cm}|P{1.2cm}}
			\multirow{2}{*}{Stage} &  \multicolumn{2}{c|}{Filters} & \multicolumn{2}{c}{Output size $T$\x$S^\text{2}$} \\
			& \multicolumn{1}{c|}{Coarse} & Fine & Coarse & Fine \\
			\shline
			\multirow{1}{*}{input} & \multicolumn{2}{c|}{\multirow{1}{*}{stride 10, 1$^\text{2}$}}  &  $T$\x224$^\text{2}$ & $T^{'}$\x224$^\text{2}$\\
			\hline
			\multirow{1}{*}{conv$_1$} & \multicolumn{2}{c|}{\multirow{1}{*}{1\x3$^\text{2}$, 3\x1, {24}}}  &  $T$\x112$^\text{2}$  & $T^{'}$\x112$^\text{2}$  \\
			\hline
			\multirow{3}{*}{res$_2$}  & \multicolumn{2}{c|}{\blocket{{54}}{{54}}{{24}}{3}} & \multirow{3}{*}{$T$\x56$^\text{2}$} & \multirow{3}{*}{$T^{'}$\x56$^\text{2}$}  \\
			& & & & \\
			& & & & \\
			\hline
			\multirow{1}{*}{grid pool$\;$} & \multirow{1}{*}{stride 1/\gridcolor{$\alpha$}, 1$^\text{2}$} & \multicolumn{1}{|c|}{\multirow{1}{*}{-}} & \gridcolor{$\alpha$}$T$\x56$^\text{2}$ & $T^{'}$\x56$^\text{2}$ \\
			\hline
			\multirow{3}{*}{res$_3$}  & \multicolumn{2}{c|}{\blocket{{108}}{{108}}{{48}}{5}} & \multirow{3}{*}{\gridcolor{$\alpha$}$T$\x28$^\text{2}$} & \multirow{3}{*}{$T^{'}$\x28$^\text{2}$} \\
			& & & & \\
			& & & & \\
			\hline
			\multirow{3}{*}{res$_4$}  & \multicolumn{2}{c|}{\blocket{{216}}{{216}}{{96}}{11}} & \multirow{3}{*}{\gridcolor{$\alpha$}$T$\x14$^\text{2}$} & \multirow{3}{*}{$T^{'}$\x14$^\text{2}$}  \\
			& & & & \\
			& & & & \\
			\hline
			\multirow{3}{*}{res$_5$}  & \multicolumn{2}{c|}{\blocket{{432}}{{432}}{{192}}{7}} & \multirow{3}{*}{\gridcolor{$\alpha$}$T$\x7$^\text{2}$} & \multirow{3}{*}{$T^{'}$\x7$^\text{2}$}  \\
			& & & & \\
			& & & & \\
			\hline
			\multirow{1}{*}{conv$_5$} & \multicolumn{2}{c|}{\multirow{1}{*}{1\x1$^\text{2}$, {432}}}   &  \gridcolor{$\alpha$}$T$\x7$^\text{2}$  & $T^{'}$\x7$^\text{2}$ \\
			\hline
			\multirow{1}{*}{pool$_5$} & \multicolumn{2}{c|}{\texttt{None}\x7$^\text{2}$}   & \multirow{3}{*}{\gridcolor{$\alpha$}$T$\x1$^\text{2}$}  & \multirow{3}{*}{$T^{'}$\x1$^\text{2}$} \\
			\multirow{1}{*}{fc$_1$} & \multicolumn{2}{c|}{\multirow{1}{*}{1\x1$^\text{2}$, {2048}}}    &   & \\
			\multicolumn{1}{c|}{ fc$_2$}  & \multicolumn{2}{c|}{\multirow{1}{*}{1\x1$^\text{2}$, { \#classes}}} & & \\
			\hline
			\multirow{1}{*}{grid unpool$\;$} & \multirow{1}{*}{stride \gridcolor{$\alpha$}, 1$^\text{2}$} & \multicolumn{1}{|c|}{\multirow{1}{*}{-}} & $T$\x1$^\text{2}$ & $T^{'}$\x1$^\text{2}$ \\
	\end{tabular}}
	
	\vspace{3mm}
	\caption{\textbf{Coarse-Fine Network Architecture} is adopted from X3D \cite{feichtenhofer2020x3d}, more specifically from the version X3D-M. Both streams have the same design and parameters, except for the addition of Grid Pool layer and Grid Unpool operation in the Coarse stream. The Fine stream process the entire temporal length of the input $T^{'}$ to provide a fine-grained context, whereas the Coarse stream can look at a segmented clip of length $T$, for which the frame-wise predictions are required. Here, $\alpha<1$ and $\alpha T$ is an integer. The kernel shapes follow the standard notation $\{T\times S^{2}, C\}$.}
	\label{tab:arch}
	\vspace{-5mm}
\end{table}

The network architecture used as the backbone in Coarse-Fine Networks is adopted from X3D \cite{feichtenhofer2020x3d}, which follow a ResNet \cite{he2016deep} structure, but designed for efficiency in video models. Both Coarse and Fine streams are initialized with separate sets of parameters, but have the same number of layers and filters as shown in \tref{tab:arch}, except for the addition of Grid Pool in the Coarse stream. Since the X3D architecture perform no temporal downsampling or pooling, it follows aggressive striding at the input level to be computationally efficient, which is set be a stride of $10$ in our case. This allows the input of the Coarse stream to cover a large temporal region, compared to what common backbones such as I3D \cite{carreira2017quo} cover during training. This is beneficial, particularly in datasets with longer temporal duration. The Coarse stream takes in segmented clips of $T=64$ frames to follow the standard X3D architecture after the Grid Pool layer (with $\alpha=1/4$) during training, while processing the input fully convolutionally at inference (i.e., $T=128$ frames during testing). In contrast, the Fine stream always process the entire input clip, which is capped at a maximum of $T^{'}=128$ frames. This limit should be adjusted based on the dataset to include the entire duration of input clips. We found $T^{'}=128$ frames with a stride of $10$ is adequate to cover the entire temporal length of more than $90\%$ of the Charades \cite{sigurdsson2016hollywood} videos.

The main difference between the Coarse stream and the Fine stream is the Grid Pool layer and the corresponding Grid Unpool operation. We want to perform this learned temporal downsampling as early as possibly in the network to reduce the compute, but at the same time, having good enough features to learn the grid locations. Thus, we place the Grid Pool layer after the first residual block $\texttt{res}_\texttt{2}$. We find that downsampling by a factor of $4$ works well in practice, to have a good compute/performance trade-off (\tref{tab:ablation:gridpool}). To calculate the confidence values ($p$) in the Grid Pool layer, we use a lightweight head ($h$) of $3$ strided convolutions with a total temporal stride of $4$ and a spatial stride of $8$, followed by spatial average pooling and a sigmoid non-linearity. The Grid Unpool operation has no learnable parameters. It is coupled with the grid locations predicted by the Grid Pool layer to perform the inverse operation of the former to recover the original temporal structure at the logits level.

We try to follow a lightweight design in Multi-stage Fusion as well. The self-attention mask $\mathbf{\hat{X}}^F_{l_i}$ is calculated through a head ($g$) of $2$ point-wise (i.e. $1\times1\times1$) convolutions followed by a sigmoid non-linearity. The Coarse-centric Gaussians ($G^C$) have no learnable parameters, and the peak magnitude of each mask is normalized to $1$. The standard deviation $\sigma$ is set to be $T^{'}/8$, empirically. The two heads $f^{\mathbf{A}_{l_i}}$ and $f^{\mathbf{A}_{l_i}}$ at each fusion location which project the concatenated multi-stage features ($\mathbf{\hat{X}}^F$) to scale ($\mathbf{A}_{l_i}$) and shift ($\mathbf{B}_{l_i}$) features contain a single point-wise convolution each. The scale features go through an additional sigmoid non-linearity. We further discuss the complexity (compute and parameters) of these operations in our ablations (subsection \ref{subsubsec:ablations}).

\vspace{-2mm}

\section{Experiments}
\vspace{-1mm}

We evaluate Coarse-Fine Networks on two large-scale benchmarks for activity detection: Charades \cite{sigurdsson2016hollywood} and MultiTHUMOS \cite{yeung2018every}. Note that we focus on the temporal detection (i.e., localization) task of generating multi-label activity annotations at every time step, which is more challenging than video classification. Activities may temporally overlap (e.g., sitting and eating), and the model must be trained to annotate all of them at each time step.

\vspace{-1mm}
\subsection{Charades}
\label{subsec:charades}
\vspace{-1mm}

\paragraph{Dataset:} Charades \cite{sigurdsson2016hollywood} is a large scale dataset consisting of $\app$9.8k continuous videos with frame-wise annotations of 157 common household activities. The dataset is split as $\app$7.9k training and $\app$1.8k validation videos. Each video contains an average of 6.8 activity instances, often with multiple activity classes per frame, and has longer clips averaging a duration of $\app$30 seconds. Such a long duration makes it a suitable dataset to test Coarse-Fine Networks.

\vspace{-3pt}
\paragraph{Training:} We initialize both Coarse and Fine streams of our network with the X3D backbone pretrained on Kinetics-400 \cite{kay2017kinetics}. For the actual training of the Coarse-Fine network as well as baselines, we follow a two-stage training process: first, training the two streams separately, followed by finetuning the combined streams. 

In the first stage, the Coarse stream considers an input of 64 frames sampled at a stride of 10, whereas the Fine stream considers 16 frames sampled at the same stride. This allows both streams to process same-sized features after Grid Pool layer. We use $\alpha=1/4$ in our experiments. Each stream is trained for 100 epochs with a batch size of 16 at a learning rate of 0.02 at the start, which is decreased by a factor of 10 at 60 and 80 epochs.

In the second stage, the two streams are trained together as Coarse-Fine Networks, with Multi-stage Fusion parameters initialized from scratch. We train this for another 100 epochs with the same schedule and batch size, but use 10$\times$ increased learning rate for the newly-initialized parameters of the fusion layers. Here, the Fine stream process the entire duration of the input, which is capped at 128 frames (sampled at a stride of 10) for Charades \cite{sigurdsson2016hollywood}. During both stages, each input is randomly sampled in [256,320] pixels, spatially cropped to 224$\times$224 and applied a random horizontal flip. We use a dropout rate of $0.5$ before the logits layer. The logits go through a sigmoid to make multi-label predictions for each frame. We use an average of classification and localization loss for training, similar to previous methods \cite{piergiovanni2018learning,piergiovanni2019temporal}.

\vspace{-3pt}
\paragraph{Inference:} At inference, we make predictions for every frame. Even though our input is sampled at a stride of 10, we consider the labels for all frames (at a stride of 1) and interpolate the logits to fit the original temporal length. In other words, we evaluate our models so that the predictions are more fine-grained at original temporal resolution. However, all state-of-the-art methods in \tref{tab:charades} report the performance for 25 equally-sampled frames per each input, which is the original Charades localization evaluation \cite{sigurdsson2016hollywood} setting. Therefore, to make a fair comparison, we evaluate our models in the same setting, only making predictions for 25 equally-sampled frames per input. The evaluation script from the Charades challenge scales the Average Precision for each class with a corresponding class weights.
At inference, the inputs are center-cropped to 224$\times$224. We report the performance as mean Average Precision (mAP) and measure the compute requirement to process an input clip of 128$\times$10 frames, for which our network processes only 128 frames due to input striding. The compute is reported as GFLOPs (floating-point operations $\times 10^9$) and the number of learnable parameters in millions (M), i.e., $\times 10^6$. We do not take advantage of any multi-crop inference.

\vspace{-3pt}
\paragraph{Results:}
\label{subsubsec:main_results}

\begin{table}[t!]
	\vspace{-6pt}
	\hspace*{-7pt}
	\centering
	\tablestyle{1.8pt}{1.05}
	\resizebox{1.0\linewidth}{!}{
		\begin{tabular}{l|c|c|c|r|r}
			\multicolumn{1}{c|}{model}  & flow & obj.  & mAP (\%) & {\scriptsize GFLOPs}  & Param (M) \\
			\shline
			{I3D (Inception) \cite{carreira2017quo}} & {} & {} & 15.63 & 2223.03 & 12.45  \\ 
			Two-stream I3D \cite{carreira2017quo} & \checkmark & {} & 17.22 & 4446.10 & 24.90 \\ 
			3D ResNet-50 \cite{tran2018closer,he2016deep} & {} & {} &   18.60 & 3187.63 & 46.52 \\ 
			{X3D \cite{feichtenhofer2020x3d}} & {} & {} &   18.87 & 37.96 & 3.29 \\
			{X3D-L \cite{feichtenhofer2020x3d}} & {} & {} &   20.03 & 147.04 & 5.78 \\
			\hline
			{I3D + super-events \cite{piergiovanni2018learning}} & \checkmark & {} & 19.41 & 4446.15 & 26.18  \\ 
			{I3D + TGM \cite{piergiovanni2019temporal}} & \checkmark & {} & 21.50 & 4446.66 & 27.00 \\
			{I3D + super-events + TGM \cite{piergiovanni2019temporal}} & \checkmark & {} & 22.30 & 4446.75 & 28.28 \\ 
			{I3D + STGCN \cite{ghosh2020stacked}} & \checkmark & \checkmark & 19.09 & 4450.94 & 29.18 \\ 
			{I3D + biGRU + VS-ST-MPNN \cite{mavroudi2020representation}} & {} & \checkmark & 23.70 & 2223.03+ & 12.45+ \\ 
			{X3D \cite{feichtenhofer2020x3d} + TGM \cite{piergiovanni2019temporal}} & {} & {} &  20.01 & 38.26 & 4.35 \\
			\hline
			SlowFast$_{\text{det}}$ (with X3D) & {} & {} & 22.31 & 54.31 & 7.41  \\
			{\textbf{Fine-Fine} (ours)} & {} & {} & \textbf{24.43} & 94.80 & 7.80  \\
			{\textbf{Coarse-Fine} (ours)} & {} & {} & \textbf{25.10} & 73.37 & 7.82  \\

	\end{tabular}}
	\vspace{1mm}
	\caption{\textbf{Comparison with the state-of-the-art methods for activity detection on Charades} \cite{sigurdsson2016hollywood}. We report the performance (mAP), compute requirement to process a clip of $128\times10$ frames (GFLOPs), and the number of parameters (M). These results correspond to the original Charades localization evaluation settings. Coarse-Fine Networks significantly outperform the previous state-of-the-art with $+1.4\%$mAP relative improvement, while reducing the compute requirement by more than one order of magnitude. It is worth noting that we do not use additional input modalities, i.e., optical-flow or object detections as any of the previous methods. The source of \cite{mavroudi2020representation} is not available to us to calculate its exact complexity values.}
	\vspace{-4mm}
	\label{tab:charades}
\end{table}

We compare the performance of the Coarse-Fine Networks with the state-of-the-art methods on Charades \cite{sigurdsson2016hollywood} localization task (i.e., temporal activity detection) in \tref{tab:charades}. For this evaluation, we use the standard test setting (i.e., the official `Charades\_v1\_localize' evaluation) where we make predictions for equally-sampled 25 frames in the validation set. This is the same procedure followed in previous work \cite{piergiovanni2018learning, piergiovanni2019temporal, mavroudi2020representation}. We report the performance (mAP), compute requirement to process a clip of $128\times10$ frames (GFLOPs) and the number of parameters (M). 

We are able to confirm that our Coarse-Fine Network performs better than all previous methods, establishing the new state-of-the-art of $25.10\%$ mAP on Charades localization. The Coarse-Fine Network, which only uses RGB, is not only superior to the previous RGB models but also is better than the methods using additional inputs modalities (i.e. optical-flow and object detection). It shows a relative improvement of $+1.4\%$ mAP compared to the previous best performing method \cite{mavroudi2020representation}, which benefits from additional training data (for its object detector training) and additional input modality (objects). 

We also note that the Coarse-Fine Network is extremely computationally efficient. Compare to the previous models,
it often requires only $\app$1$/$75 of computations (e.g., 73 vs. 4446 GFLOPS). Further, this computation is without considering the overhead for optical flow computation or object detection in prior works. 
The significant computation efficiency of the Coarse-Fine Networks is due to the better utilization of the RGB features, which discards the need for processing additional modalities, as well as an efficient usage of X3D modules with our temporal Grid Pooling and Multi-stage Fusion, which we confirm further with our ablations in the next section.



We further report a version of our method: Fine-Fine Networks, in which the Grid Pool layer is removed from the Coarse stream, to highlight the importance of the Coarse features. Fine-Fine Networks still benefit from our Multi-stage Fusion. The Grid Pool operation dynamically sample important frames to generating a coarse temporal resolution, which gives the Coarse-Fine Networks an relative performance gain of $+0.67\%$ mAP and $23\%$ computation reduction. We also evaluated the baseline extension of X3D as a two-stream network (with different temporal resolutions) in a form similar to \cite{feichtenhofer2019slowfast}, which we name SlowFast$_{\text{det}}$. This does not have our Grid Pool layer or the Multi-stage Fusion mechanism. The result shows the benefits of the components, giving a relative improvement of $2.79\%$mAP. A larger version of X3D (i.e., X3D-L) shows that the performance improvement of Coarse-Fine compared to X3D is not merely due to the increased compute.

It is important to also note that all previous methods work on pre-extracted features from a frozen backbone, essentially making them late-modeling techniques, either using graph-based methods \cite{ghosh2020stacked,mavroudi2020representation} or abstracting long-term temporal information \cite{piergiovanni2018learning, piergiovanni2019temporal}. In contrast, our method allows end-to-end training of feature fusions at intermediate locations of the network, enabling it to learn better representations using only RGB information.

\fref{fig:performance_complexity} further highlights the benefit of Coarse-Fine Networks compared to previous state-of-the-arts. We show the performance/complexity trade here, with the x-axis (complexity in GFLOPs) in log-scale. Our method shows comparable performance with the previous best performing method which outperforms all previous state-of-the art methods, while being extremely efficient in design. 

\vspace{-1mm}
\subsection{Ablations}
\label{subsubsec:ablations}
\vspace{-1mm}

\begin{table*}[t!]\centering
        \vspace{-3em}
		\captionsetup[subfloat]{captionskip=2pt}
		\captionsetup[subffloat]{justification=centering}
		\subfloat[\textbf{Fusion location}: Using the fusion connections only before the logits, in-between each \texttt{res} block w/ or w/o considering multiple abstraction-levels at each fusion location. (Fine-Fine)
		\label{tab:ablation:fuse_loc}]{
			\tablestyle{2pt}{1.05}
			\begin{tabular}{l|c|r}
			  \multicolumn{1}{c|}{Fusion location}  & mAP (\%) & {\scriptsize GFLOPs} \\
			\shline
			 late only& 21.84 & 77.15 \\ 
			 late$+$intermid one-to-one & 22.50 & 81.80 \\
			 late$+$intermid multi-stage & 22.65 & 94.80 \\
			 \multicolumn{3}{c}{} \\ 

	        \end{tabular}}\hspace{3mm}
		\subfloat[\textbf{Fusion dimensions}: The Dimensions of Multi-stage Fusion features. When temporal dimension (T) is available, we use Coarse-centric Gaussians for location calibration. (Fine-Fine) \vspace*{-3mm}
		\label{tab:ablation:fuse_dim}]{
			\tablestyle{2pt}{1.05}
			\begin{tabular}{l|c|r}
			 \multicolumn{1}{c|}{Fusion dimensions}  & mAP (\%) & {\scriptsize GFLOPs} \\
			\shline
			  C & 18.11 & 76.45 \\ 
			  CHW & 19.86 & 93.12 \\
			  CTHW & 22.65 & 94.80 \\
			  \multicolumn{3}{c}{} \\ 

	        \end{tabular}}\hspace{3mm}
		\subfloat[\textbf{Fusion mask}: The effect of using a self-attention mask to filter the fine-grained context (refer \fref{fig:fusing}), with different fusion connections. (Fine-Fine)
		\label{tab:ablation:attn}]{
			\tablestyle{2pt}{1.05}
			\begin{tabular}{l|l|c|r}
			 \multicolumn{1}{c|}{Fusion loc.} & \multicolumn{1}{c|}{Fusion mask}  & mAP (\%) & {\scriptsize GFLOPs} \\
			\shline
			  \multirow{2}{*}{late} & none & 20.59 & 75.98 \\ 
			  & self-attention & 21.84 & 77.15 \\ \hline
			  \multirow{2}{*}{multi-stage} & none & 21.42 & 92.69 \\
			  & self-attention & 22.65 & 94.80 \\ 

	        \end{tabular}}\hspace{3mm}
		\subfloat[\textbf{Pooling type}: Different types of temporal pooling operations used after $\texttt{res}_{\texttt{2}}$ block. A temporal stride of 4 (equivalent to $\alpha=1/4$) used here. (Coarse-only)
		\label{tab:ablation:pooling_type}]{
			\tablestyle{2pt}{1.05}
			\begin{tabular}{l|c|r}
			\multicolumn{1}{c|}{Pooling type}  & mAP (\%) & {\scriptsize GFLOPs} \\
			\shline
			Max & 16.21 & 15.42 \\ 
			Average & 16.64 & 15.42 \\
			Striding & 17.49 & 15.42 \\
			Grid Pool & 18.12 & 16.53 \\

	        \end{tabular}}\hspace{3mm}
		\subfloat[\textbf{Grid Pool configuration}: Variations of the sampling rate $\alpha$ with different temporal sizes of the input at the Grid Pool layer. (Coarse-only)
		\label{tab:ablation:gridpool}]{
			\tablestyle{2pt}{1.05}
			\begin{tabular}{l|c|c|r}
			\multicolumn{1}{c|}{Grid Pool input} & $\alpha$ & mAP (\%) & {\scriptsize GFLOPs} \\
			\shline
			\multirow{2}{*}{$T\hspace{-1mm}=\hspace{-1mm}128$, $\text{stride}\hspace{-0.5mm}=\hspace{-1mm}10$} & 1/4 & 18.12 & 16.53 \\ 
			& 1/8 & 11.88 & 10.43 \\ 
			\hline
			\multirow{2}{*}{$T\hspace{-1mm}=\hspace{-1mm}256$, $\text{stride}\hspace{-0.5mm}=\hspace{-0.5mm}5$} & 1/4 & 18.16 & 32.82 \\ 
			& 1/8 & 15.56 & 20.63 \\

	        \end{tabular}}\hspace{3mm}
		\subfloat[\textbf{Importance of Grid Pool and Multi-stage Fusion combined}: SlowFast$_\text{det}$ is a baseline w/o Grid Pool and Multi-stage Fusion. It samples input at different frame-rates (Fast is $\times4$ faster) and uses fusion connections similar to that of SlowFast \cite{feichtenhofer2019slowfast}. (Coarse-Fine) \label{tab:ablation:slowfast}]{
			\tablestyle{2pt}{1.05}
			\begin{tabular}{l|c|r}
			\multicolumn{1}{c|}{Two-stream Network}  & mAP (\%) & {\scriptsize GFLOPs} \\
			\shline
			SlowFast$_{\text{det}}$ & 20.61 & 54.31 \\ 
			SlowFast$_{\text{det}}$ (w/ Grid Pool) & 20.82 & 55.42 \\
			SlowFast$_{\text{det}}$ (w/ Fusion) & 22.79 & 72.16 \\
		    \textbf{Coarse-Fine} (w/ Grid Pool w/ Fusion)   & 23.24 & 73.37 \\
			\end{tabular}}
		
		\vspace{0.5em}
		\caption{\textbf{Ablations on Charades} \cite{sigurdsson2016hollywood} \textbf{localization} comparing the design choices of Grid Pool and Multi-stage Fusion. We show the performance in mean Average Precision (mAP), and measure the compute requirement for a temporal clip of $T\hspace{-1mm}=128$ at inference in GFLOPs (floating-point operations $\times 10^9$). In these tables, we report the performance for fine-grained predictions, making decisions for every frame. The network configuration used in each experiment is shown within the braces of each caption (Fine-Fine, Coarse only or Coarse-Fine). Fine-Fine Networks are used in Fusion experiments to decouple the effect of Grid Pool, and similarly, Coarse only Networks are used in Grid Pool experiments decouple the effect of Multi-stage Fusion.}
		\label{tab:ablations}
		\vspace{-4mm}
\end{table*}
	
Here, we discuss multiple ablation experiments validating our design decisions, specifically on our Grid Pool layer and Multi-stage Fusion. We use the Charades dataset (with the localization setting as above).

In our ablation experiments, we take advantage of more robust evaluation metric to compare our approach and the baselines. We make the model to generate a multi-class activity annotation at every time step, and compare it with the ground truth to measure the mAP. This is very similar to the official `Charades\_v1\_localize setting' used above, except that (i) it is evaluated for $\times$25 more frames for the completeness and that (ii) we measure mAP per activity class and average them.

\vspace{-3pt}
\paragraph{Fusion location:} First, we explore which locations in our two stream architecture would be ideal to implement the fusion connections. We consider the late fusion as a baseline, in which, the only fusion happens just before the logits layer. This is similar to the previous methods in \cite{piergiovanni2018learning, piergiovanni2019temporal}. Next, We extend this fusion to multiple intermediate levels, specifically, after each \texttt{res} block, in which we fuse the two streams at only equivalent abstraction levels, i.e., at the same depth. This is similar to the fusion in SlowFast \cite{feichtenhofer2019slowfast}. Finally, we consider multiple abstraction-levels of Fine features for the fusion, which gives our Multi-stage Fusion. The results of this ablation is given in \tref{tab:ablation:fuse_loc}. Note that here we evaluate our fusion in a Fine-Fine Network to decouple the effect of Grid Pool from our fusion. Multi-stage Fusion shows a $+0.81\%$ mAP improvement compared to only using a late fusion. The improvement of considering multiple abstraction-levels is marginal, at $+0.15\%$ mAP, suggesting that features at the same abstraction level can provide the most complementary information.

\vspace{-3pt}
\paragraph{Fusion dimensions:} We experiment on the significance of different dimensions in the fusion features. Either having only channel dimension (C) similar to \cite{piergiovanni2018learning}, channel-spatial (CHW) dimensions or all channel-temporal-spatial (CHWT) similar to \cite{feichtenhofer2019slowfast} is considered here. The results of this experiment is reported in \tref{tab:ablation:fuse_dim}. Note that the dimensions which are not available in any of the above scenarios are average pooled before the fusion. We see that having all CHWT dimensions in the fusion feature has a large improvement compared to the baseline, specifically $+4.54\%$ mAP. Introduction of the temporal dimension (T) shows the most improvement, which is $+2.79\%$ mAP. This is in fact mainly due to the temporal Gaussians in our Fusion that calibrate the features based on the location, without which, we can not see such improvement (i.e., +0.61\% mAP over a single stream, when naively selecting corresponding temporal locations in the two streams for fusion w/o Gaussians).

\vspace{-3pt}
\paragraph{Fusion mask:} Here, we evaluate how important it is to filter the Fine features at the input of fusion, results of which, is shown in \tref{tab:ablation:attn}. In the Multi-stage Fusion setting, having a self-attention mask to adaptively weight each Feature point gives an improvement of $+1.23\%$ mAP compared to feeding the Fine feature directly.

\vspace{-3pt}
\paragraph{Pooling type:} Next, we explore the performance gain caused by the proposed (temporal) Grid Pool layer. We compare against conventional temporal pooling operations such as max pooling, average pooling and even simple temporal striding. Here, we report the numbers for a Coarse-only network to decouple the effect of Grid Pool from that of Multi-stage Fusion. In these experiments, we set $\alpha=1/4$, which essentially means a $\times$4 temporal downsampling, and perform the downsampling after the \texttt{res}$_{\texttt{2}}$ block. Max pooling, average pooling use a similar setting of kernel size of 4 and a stride of 4. Grid pooling gives a consistent improvement over other methods, specifically $+1.91\%$ mAP and $+1.48\%$ mAP over commonly used max pooling and average pooling, respectively. We also note that a simple temporal striding can outperform max pooling and average pooling by $+1.28\%$ mAP and $+0.85\%$ mAP, respectively. We suspect that the inferior performance of max/average pooling is due to the aggressive temporal striding at the input of X3D, which is a stride of 10 by default (i.e., after the pooling, the stride is 40). In such a large window, pooling tends to blur most of the temporal information.

\vspace{-3pt}
\paragraph{Grid Pool configuration:} We try different configurations of Grid Pooling to evaluate its performance and compute requirement. Similar to above, we use the Coarse-only network. We consider an input of temporal length $T=128$  at a stride of 10 or $T=256$ at a stride of 5, to cover entire duration of Charades videos \cite{sigurdsson2016hollywood}. We try temporally downsampling each of these with $\alpha=1/4$ ($\times$4 downsampling) or $\alpha=1/8$ ($\times$8 downsampling). The performance of these configurations is given in \tref{tab:ablation:gridpool}. $\times$8 downsampling shows a significantly lower performance indicating that it looses too much information with such an aggressive stride, i.e., more frames are important and need to be sampled by the Grid Pool layer. Moreover, increasing the number of input frames does not necessarily improve the performance (only $+0.02\%$ mAP) with $\alpha=1/4$. Among these configurations, $T=128$ with $\alpha=1/4$ shows the best performance/compute trade-off.

\vspace{-3pt}
\paragraph{Grid Pool and Multi-stage Fusion combined:} Finally, we evaluate the combined performance of Grid Pool and Multi-Stage Fusion. To do this, we consider a two-stream baseline without these components, which we call SlowFast$_\text{det}$. This performs $\times$4 temporal downsampling in the Coarse stream based on striding, and use direct frame correspondences between Fine and Coarse streams for fusion, similar to SlowFast \cite{feichtenhofer2019slowfast} 
while still using X3D modules like ours. The results of this study is given in \tref{tab:ablation:slowfast}. When compared with this baseline, introduction of either Grid Pool or Multi-stage Fusion provides consistent improvements of $+0.21\%$ mAP and $+2.18\%$ mAP respectively. Our Coarse-Fine Networks outperform this baseline by a margin of $+2.63\%$ mAP.

\vspace{-3pt}
\paragraph{Trade-off with X3D:} Coarse-Fine network is designed to use a similar amount of computation as a two-stream version of X3D-M. Another way to use the extra compute is by increasing the number of layers. To understand if the increased compute is meaningful, we test X3D-L, a larger version of X3D (\tref{tab:charades}). X3D-L shows of $20.03\%$ mAP with a compute of 147.04 GFLOPS. Coarse-Fine Networks outperform this in both accuracy and speed with $25.10\%$ mAP at 73.37 GFLOPS.

\vspace{-1mm}
\subsection{MultiTHUMOS}
\label{subsec:multithumos}
\vspace{-1mm}

\paragraph{Dataset:} MultiTHUMOS \cite{yeung2018every} is an extension of the THUMOS \cite{jiang2014thumos} dataset with the untrimmed videos densely annotated for 65 different action classes. It provides frame-level action annotations for 30 hours of video across 413 videos, split as 200 for training and 213 for validation. On average, it contains 1.5 labels per frame and 10.5 action classes per video. It contains significantly smaller number of videos compared to Charades \cite{sigurdsson2016hollywood} and each video has a larger temporal length, which make the training difficult. We created a segmented version of this data, where each clip is limited to a maximum of 1280 frames, which gives a similar evaluation setting to Charades. For the computational efficiency, we use the temporal striding of 10. 

\vspace{-3pt}
\paragraph{Training:} We follow a training process similar to what we did for Charades. We follow two stages of training with our Coarse and Fine streams pretrained on Kinetics-400 \cite{kay2017kinetics}, i.e., separately and combined. We use the same hyperparameter settings and training schedules as in Charades (refer subsection \ref{subsec:charades}). We use a dropout rate of $0.5$ before the logits layer. The logits go through a sigmoid to make multi-label predictions for each frame. We use an average of classification and localization loss for training.

\vspace{-3pt}
\paragraph{Inference:} At inference, we make predictions for every frame. Even though our input is sampled at a stride of 10, we consider the labels for all frames (at a stride of 1) and interpolate the logits to fit the original temporal length. Each input is center-cropped to 224$\times$224. We report the performance (mAP), compute requirement to process an input clip of 1024$\times$10 frames as TFLOPs ($\times 10^9$) and the number of learnable parameters(M). The length of 1024$\times$10 frames is only considered as a reference for reporting the complexity values, and there are longer clips in the dataset with up to $\times$5 frames. We process these frames fully convolutionally.

\vspace{-3pt}
\paragraph{Results:}
\label{subsubsec:main_results_thumos}

\begin{figure}[t]
	\centering
	\vspace{-2mm}
	\includegraphics[width=0.9\linewidth]{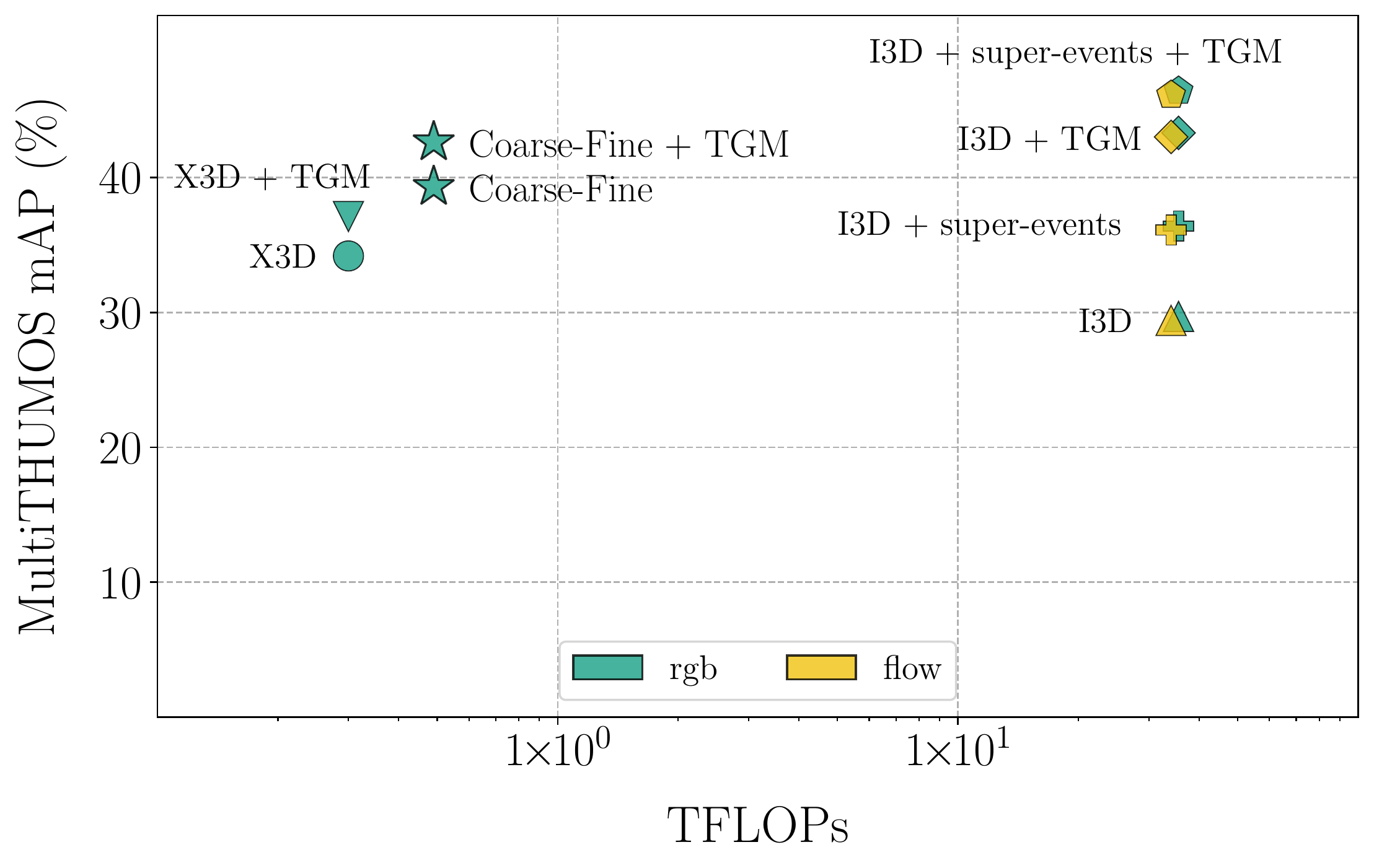}
	\vspace{-1mm}
	\caption{\textbf{Performance/complexity trade-off} of state-of-the-art methods for activity detection on MultiTHUMOS \cite{yeung2018every}. Our Coarse-Fine Networks /w TGM show a comparable performance with the state-of-the-arts with $\sim$75x speed, without using additional input modalities.}
	\label{fig:complexity_thumos}
	\vspace{-5mm}
\end{figure}

We show the performance (mAP) of Coarse-Fine Networks on MultiTHUMOS \cite{yeung2018every} activity detection with the corresponding compute requirement (TFLOPs, i.e., $\times 10^{12}$) in \fref{fig:complexity_thumos}. We observe an improvement of $+4.63\%$ from X3D \cite{feichtenhofer2020x3d} to Coarse-Fine. While our Coarse-Fine network is almost 75 times faster than the previous model (0.49 TFLOPS (Coarse-Fine) vs. 35.57 TFLOPS (I3D + TGM)), it still achieves comparable performance to the previous state-of-the-art. Models using the X3D backbone including ours lose motion details due to the aggressive 1/10 striding compared to I3D \cite{carreira2017quo} that doesn't do striding, making them less effective when combined with other temporal modeling methods (e.g., TGM \cite{piergiovanni2019temporal}). Still, our Corase-Fine Networks were able to overcome such limitation and perform comparably. Coarse-Fine /w TGM shows a further improvement of $+2.21\%$mAP.



\vspace{-2mm}
\section{Conclusion}
\vspace{-2mm}

We presented Coarse-Fine Networks, which is a two-stream architecture to combine temporally Coarse representations with Fine representations.
We introduced the approach of temporal Grid Pooling that learns to differentiably select informative frames while discarding the other, to obtain Coarse representations. We also introduced the Multi-stage Fusion to best combine the Coarse stream with the Fine stream. We confirmed that the Coarse-Fine networks obtain the best known performance on Charades localization, while spending much less computation time.

\paragraph{Acknowledgements:} This work was supported by the National Science Foundation (IIS-2104404 and CNS-2104416). The authors thank AJ Piergiovanni for helpful discussions.

{\small
\balance
\bibliographystyle{ieee_fullname}
\bibliography{egbib}
}

\end{document}